%% file: main.tex
\documentclass[10pt,twocolumn,letterpaper]{article}

\usepackage[pagenumbers]{cvpr} % To force page numbers, e.g. for an arXiv version
\usepackage{graphicx}
\usepackage{amsmath}
\usepackage{amssymb}
\usepackage{booktabs}
\usepackage{bm}
\usepackage{multirow}
\usepackage{caption}
\usepackage[accsupp]{axessibility}
\input{math_command.tex}

\usepackage[pagebackref,breaklinks,colorlinks]{hyperref}

\usepackage[capitalize]{cleveref}
\crefname{section}{Sec.}{Secs.}
\Crefname{section}{Section}{Sections}
\Crefname{table}{Table}{Tables}
\crefname{table}{Tab.}{Tabs.}

\def\cvprPaperID{3792} % *** Enter the CVPR Paper ID here
\def\confName{CVPR}
\def\confYear{2023}

\begin{document}

%%%%%%%%% TITLE - PLEASE UPDATE
\title{Transfer Knowledge from Head to Tail:\\
Uncertainty Calibration under Long-tailed Distribution}

\author{Jiahao Chen\textsuperscript{\rm 1 2}, Bing Su\textsuperscript{\rm 1 2 $\dag$}\\
\textsuperscript{\rm 1} Gaoling School of Artificial Intelligence, Renmin University of China\\
\textsuperscript{\rm 2} Beijing Key Laboratory of Big Data Management and Analysis Methods\\
{\tt\small \{nicelemon666, subingats\}@gmail.com}
% Institution1 address\\
% {\tt\small subingats@gmail.com  nicelemon666@gmail.com}
% For a paper whose authors are all at the same institution,
% omit the following lines up until the closing ``}''.
% Additional authors and addresses can be added with ``\and'',
% just like the second author.
% To save space, use either the email address or home page, not both
% \and
% Bing Su \textsuperscript{$\dag$}\\
% Renmin University of China\\
% First line of institution2 address\\
% {\tt\small subingats@gmail.com}
}
\maketitle

%%%%%%%%% ABSTRACT
\begin{abstract}
   How to estimate the uncertainty of a given model is a crucial problem. Current calibration techniques treat different classes equally and thus implicitly assume that the distribution of training data is balanced, but ignore the fact that real-world data often follows a long-tailed distribution. In this paper, we explore the problem of calibrating the model trained from a long-tailed distribution. Due to the difference between the imbalanced training distribution and balanced test distribution, existing calibration methods such as temperature scaling can not generalize well to this problem. Specific calibration methods for domain adaptation are also not applicable because they rely on unlabeled target domain instances which are not available. Models trained from a long-tailed distribution tend to be more overconfident to head classes. To this end, we propose a novel knowledge-transferring-based calibration method by estimating the importance weights for samples of tail classes to realize long-tailed calibration. Our method models the distribution of each class as a Gaussian distribution and views the source statistics of head classes as a prior to calibrate the target distributions of tail classes. We adaptively transfer knowledge from head classes to get the target probability density of tail classes. The importance weight is estimated by the ratio of the target probability density over the source probability density. Extensive experiments on CIFAR-10-LT, MNIST-LT, CIFAR-100-LT, and ImageNet-LT datasets demonstrate the effectiveness of our method.
\end{abstract}

\input{0.introduction.tex}

\input{1.related_work.tex}
\input{2.method.tex}
\input{3.experiment.tex}
\input{4.conclusion.tex}

%%%%%%%%% REFERENCES
{\small
\bibliographystyle{ieee_fullname}
\bibliography{egbib}
}

\end{document}

%% file: math_command.tex
\usepackage{amsmath,amsfonts,bm}

\def\eqref#1{equation~\ref{#1}}
\def\1{\bm{1}}

\def\vmu{{\bm{\mu}}}

\def\vd{{\bm{d}}}

\def\vf{{\bm{f}}}

\def\vs{{\bm{s}}}

\def\vx{{\bm{x}}}

\def\vz{{\bm{z}}}

\DeclareMathAlphabet{\mathsfit}{\encodingdefault}{\sfdefault}{m}{sl}
\SetMathAlphabet{\mathsfit}{bold}{\encodingdefault}{\sfdefault}{bx}{n}

%% file: 0.introduction.tex
\section{Introduction}
\footnotetext{\textsuperscript{$\dag$}Corresponding author.}
With the development of deep neural networks, great progress has been made in image classification. In addition to performance, the uncertainty estimate of a given model is also receiving increasing attention, as the confidence of a model is expected to accurately reflect its performance. A model is called \emph{perfect calibrated} if the predictive confidence of the model represents a good approximation of its actual probability of correctness~\cite{guo2017calibration}. Model calibration is particularly important in safety-critical applications, such as autonomous driving, medical diagnosis, and robotics~\cite{amodei2016concrete}. For example, if a prediction with low confidence is more likely to be wrong, we can take countermeasures to avoid unknown risks.

\begin{figure*}[h]
    \centering
    \subfloat[Calibration under balanced distribution.]{\includegraphics[width=0.83\textwidth]{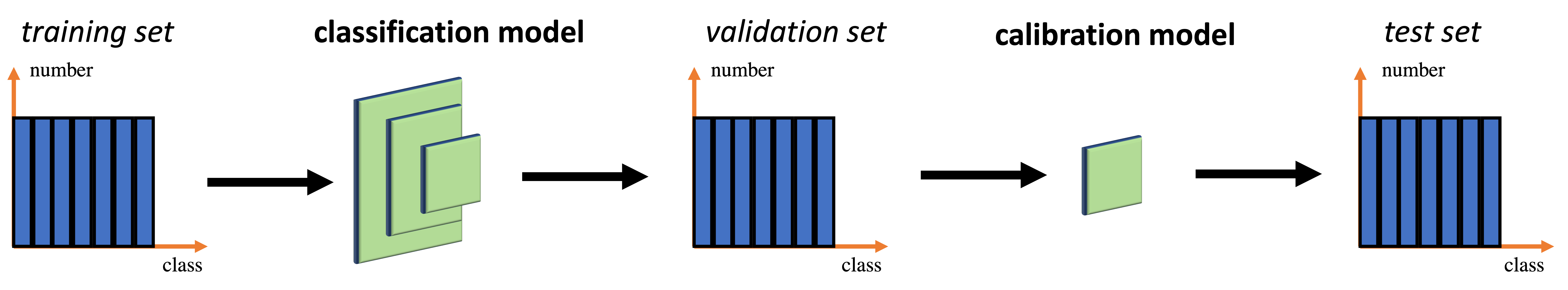}}\\
    \subfloat[Calibration under long-tailed distribution.]{\includegraphics[width=0.83\textwidth]{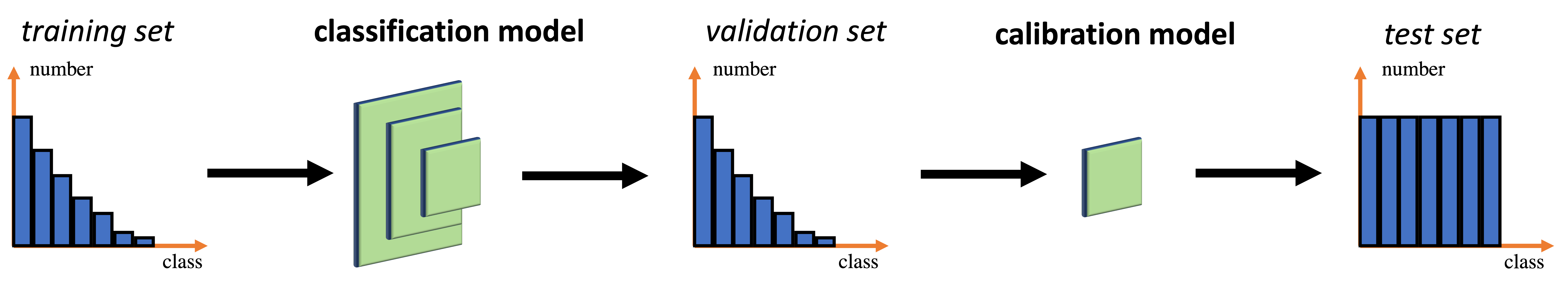}}
    \caption{The difference between calibration under balanced distribution and calibration under long-tailed distribution. (a) The classification model and the calibration model are trained on the balanced training and validation sets, respectively, and the test set is balanced. (b) The classification model and the calibration model are trained on the long-tailed training and validation sets, while the test set is balanced.}
    \label{fig:first}
    \vskip -0.15in
\end{figure*}

Most existing calibration techniques assume that the distribution of training data is balanced, i.e., each class has a similar number of training instances, so that each class is treated equally~\cite{guo2017calibration, naeini2015obtaining, kull2019dircal}. As shown in \cref{fig:first}, the traditional calibration pipeline uses a balanced training set to train the classification model and a balanced validation set to obtain the calibration model, respectively. The target test set is in the same distribution as the training/validation set. However, data in the real-world often follows a long-tailed distribution, i.e., a few dominant classes occupy most of the instances, while much fewer examples are available for most other classes~\cite{kang2020exploring, liu2019large, cui2019class}. When tested on balanced test data, classification models trained from the training set with a long-tailed distribution are naturally more over-confident to head classes. Only imbalanced validation set with the same long-tailed distribution is available for calibrating such models since the validation set is often randomly divided from the training set. 

\begin{figure}[t]
    \centering
    \subfloat[Validation set]{\includegraphics[width=0.16\textwidth]{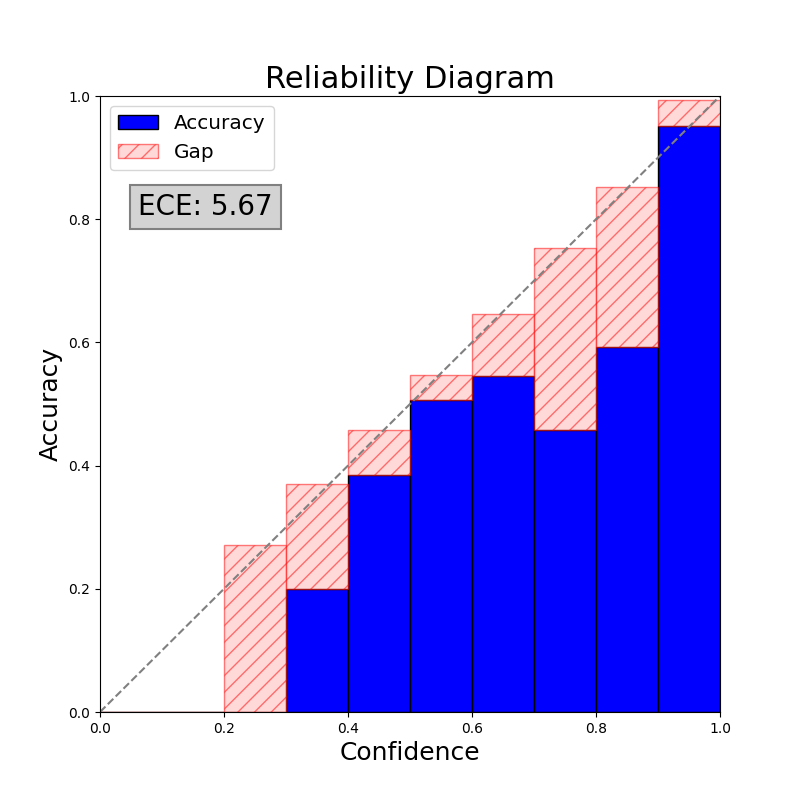}}
     \subfloat[Test set]{\includegraphics[width=0.16\textwidth]{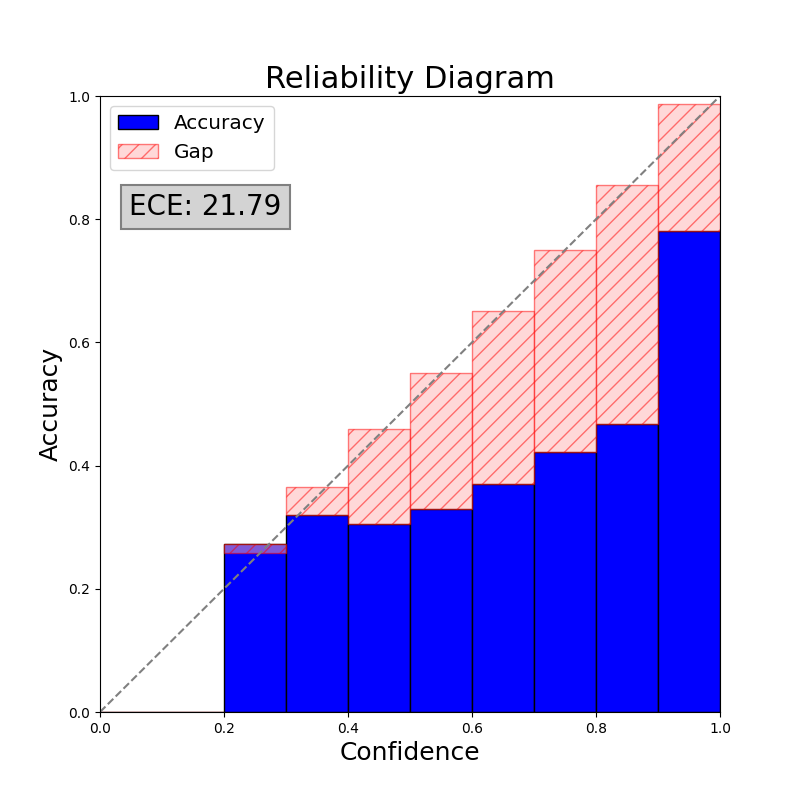}}
    \subfloat[Temperature scaling]{\includegraphics[width=0.16\textwidth]{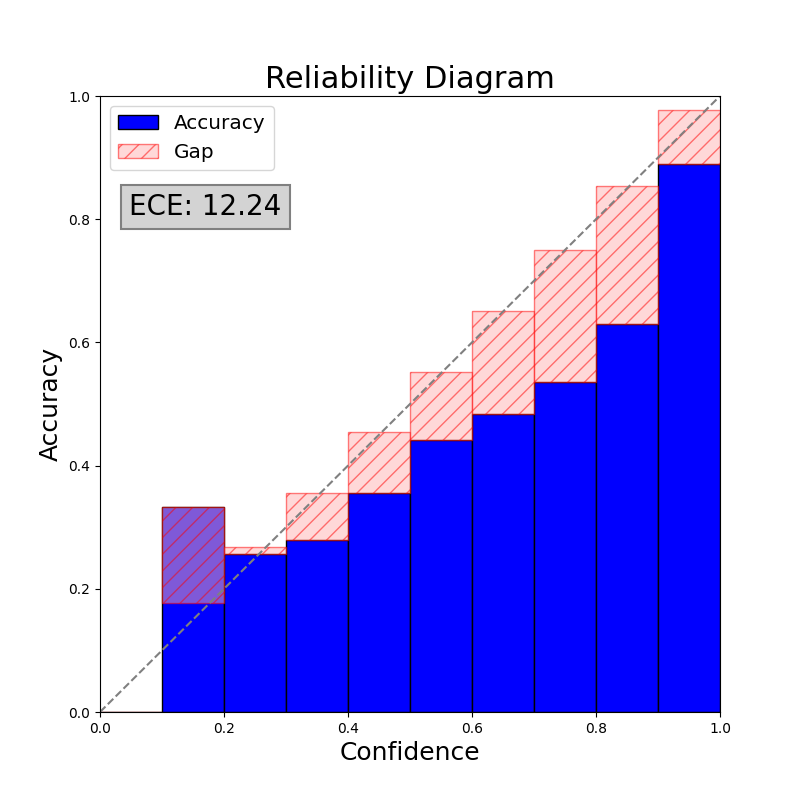}}
    \caption{The reliability diagrams of (a) the validation set before calibration, (b) the test set before calibration, and (c) the test set after calibration with temperature scaling.}
    \label{phenomenon}
    \vskip -0.1in
\end{figure}

Due to the different distributions between the imbalanced training data and the balanced test data~\cite{jamal2020rethinking}, it is difficult for traditional calibration techniques to achieve balanced calibration among head classes and tail classes with different levels of confidence estimations. For instance, temperature scaling~\cite{guo2017calibration} with the temperature learned on a validation set obtains degraded performance on the test set if the two sets are in different distribution~\cite{tomani2021post, pampari2020unsupervised}. As shown in \cref{phenomenon}, a balanced test set suffers heavier overconfidence compared with a long-tailed validation set. Although temperature scaling can relieve such phenomenon, there still exists overconfidence after calibration. Domain adaptation calibration methods~\cite{pampari2020unsupervised, wang2020transferable} aim to generalize calibration across domains under a covariate shift condition but they utilize unlabeled target domain instances. Similarly, the domain generalization calibration method~\cite{gong2021confidence} uses a support set to bridge the gap between the source domain and the target domain, which also relies on extra instances. These methods cannot be applied to the long-tailed calibration since the balanced test domain is not available.

In this paper, we investigate the problem of \textit{calibration under long-tailed distribution}. Since the distributions of each tail class in the imbalanced validation set and the balanced target set are different, we utilize the importance weight strategy to alleviate the unreliable calibration for tail classes. The weight of each instance is the ratio between the target balanced probability density and the source imbalanced probability density. We explicitly model the distribution of each class as a Gaussian distribution. Different from the source distribution, the target balanced distribution cannot be estimated directly. Since there exists common information between head classes and tail classes~\cite{liu2020deep}, we transfer knowledge from head classes to estimate the target probability density. Normally, the more similar two classes are, the more information they share. Therefore, for each tail class, we measure the similarities between the distributions of it and all head classes with the Wasserstein distance. The similarity considers both first-order and second-order statistics of the two classes and thus can better reflect the transferability of statistical knowledge. Then we estimate the target probability density of each tail class by combining its own distribution and the transferred information from all head classes referring to the similarities. Finally, we calibrate the model with the importance weights. 

Our contributions are summarized as: 1) We explore the problem of calibration under long-tailed distribution, which has important practical implications but is rarely studied. We apply the importance weight strategy to enhance the estimation of tail classes for more accurate calibration. 2) We propose an importance weight estimation method by viewing distributions of head classes as prior for distributions of tail classes. For each tail class, our method estimates its probability density function from the distribution calibrated by head classes and calculates the importance weight to realize balanced calibration. 3) We conduct extensive experiments on the CIFAR-10-LT, CIFAR-100-LT~\cite{cao2019learning}, MNIST-LT~\cite{lecun1998gradient}, ImageNet-LT~\cite{liu2019large} datasets and the results demonstrate the effectiveness of our method.

% \begin{itemize}
%     \item 1) We explore the problem of calibration under long-tailed distributions, which has important practical implications but is rarely studied. We apply the importance weight strategy to enhance the estimation of tail classes for more accurate calibration.
%     \item 2) We propose an importance weight estimation method by viewing distributions of head classes as prior for distributions of tail classes. For each tail class, our method estimates its probability density function from the distribution calibrated by head classes and calculates the importance weight to realize balanced calibration.
%     \item 3) We conduct extensive experiments on the CIFAR-10-LT, CIFAR-100-LT~\cite{cao2019learning}, MNIST-LT~\cite{lecun1998gradient}, ImageNet-LT~\cite{liu2019large} datasets and the results demonstrate the effectiveness of our method.
% \end{itemize}

%% file: 1.related_work.tex
\section{Related work}
\textbf{Post-processing calibration}. Current calibration techniques can be roughly divided into post-processing methods and regularization methods~\cite{hebbalaguppe2022stitch, cheng2022calibrating}. Post-processing methods focus on learning a re-calibration function on a given model. Platt scaling~\cite{platt1999probabilistic} transforms outputs of a classification model into a probability distribution over classes. It can solve the calibration of non-probabilistic methods like SVM~\cite{cortes1995support}. Temperature scaling~\cite{guo2017calibration} extends Platt scaling and is applied to multi-class classification problems. It optimizes a parameter $T$ to re-scale the output logits of a given model. Non-parametric isotonic regression~\cite{zadrozny2002transforming} learns a piece-wise constant function that minimizes the residual between the calibrated prediction and the labels. In~\cite{zhang2020mix}, three properties, accuracy-preserving, data-efficient, and expressive of uncertainty calibration are proposed. Experiments show that a combination of the non-parametric method and the parametric method can achieve better results. 

\textbf{Domain shift calibration}. The most common case is that the validation and test sets are in different domains~\cite{wald2021calibration}. The re-calibration function learned by the validation domain cannot be generalized to the test domain. CPCS~\cite{park2020calibrated} utilizes importance weight to correct for the shift from the training domain to the target domain and achieves good calibration for the domain adaptation model. TransCal~\cite{wang2020transferable} achieves more accurate calibration with lower bias and variance in a unified hyperparameter-free optimization framework. In~\cite{gong2021confidence}, a support set is applied to bridge the gap between the source domain and a target domain, and three calibration strategies are proposed to achieve calibration for domain generalization. In~\cite{tomani2021post}, current techniques have demonstrated that overconfidence is still existing under domain shift, and a simple strategy where perturbations are applied to samples in the validation set before performing the post-hoc calibration step is proposed.

Although long-tailed calibration also suffers the domain shift problem, it is different from domain shift calibration since unlabeled target domain instances or plenty of data to constitute a support set are not available for calibration. There are some works~\cite{zhong2021improving, xu2021towards} to improve calibration under long-tailed distribution by modifying the training objective to train new models from scratch. In contrast, our method focuses on calibrating existing models using a small validation set. Previous methods such as Gistnet~\cite{liu2021gistnet} use head classes' knowledge to enhance tail classes' generalization, but they concentrate on long-tailed classification while we address the calibration issue. 

% Previous methods such as Gistnet~\cite{liu2021gistnet} use head classes' knowledge to enhance tail classes' generalization, but they concentrate on long-tailed classification while we address the calibration issue. Although long-tailed distribution calibration also suffers the domain shift problem, it is different from domain shift calibration since unlabeled target domain instances or plenty of data to constitute a support set are not available for calibration. Therefore, we employ the importance weight and estimate the target probability density by utilizing the inherent property of long-tailed distribution. 

%% file: 2.method.tex
\section{Method}
\subsection{Notation}
We propose the problem of \emph{calibration under long-tailed distribution}. Given a long-tailed distribution $p(\vx)$ and a corresponding balanced distribution $q(\vx)$, we hold the assumption that $p(\vx) \neq q(\vx)$ while $p(y|\vx)=q(y|\vx)$. Instances are i.i.d. sampled from $p(\vx)$ to construct a long-tailed training set $\mathcal{S} = \{(\vx_i, y_i)\}$ and a validation set $\mathcal{V}$, where $y_i \in \{1, \cdots, C\}$ is the label of the $i^{th}$ instance $\vx_i$, $C$ is the number of classes, and $n_{c}$ denotes the number of instances belongs to the $c^{th}$ class. Similarly, instances are i.i.d. sampled from $q(\vx)$ to construct a balanced test set $\mathcal{T}$. Assuming classes are sorted by decreasing cardinality, i.e., $n_1 \geq n_2 \geq ... \geq n_C$, the data follows a long-tailed distribution where most instances belong to head classes, while each tail class has only a few instances. We divide all classes into head classes $\mathcal{A}_{head}=\{c | n_c \geq \zeta\}$ and tail class $\mathcal{A}_{tail}=\{c | n_c < \zeta\}$, where $\zeta$ is a threshold. Moreover, we have been given a classification model $\phi(\cdot)$ trained on $\mathcal{S}$, where the output of $\phi(\vx_i)$ is denoted by $\vz_i$ and the corresponding feature (the output of the layer before the classifier) is denoted by $\vf_i$. The goal is to calibrate the model $\phi(\cdot)$ on the validation set $\mathcal{V}$ so that the model is calibrated on the balanced test data $\mathcal{T}$.

\subsection{Background}
\textbf{Calibration}. For each instance $\vx_i$, we acquire its confidence score $\hat{p}_i$ and prediction result $\hat{y}_i$ from the output $\vz_i$. Formally, if the following \cref{definition} is satisfied, the model $\phi(\vx_i)$ is called perfect calibrated. The definitions of $\hat{p}_i$ and $\hat{y}_i$ are in \cref{definition2} and $softmax(\vz_i)$ denotes the softmax function $softmax(\vz_i) = \frac{exp(\vz_i)}{\sum_{j=1}^Cexp(\vz_j)} $.
\begin{equation}
    \mathbb{P}(\hat{y}_i=y_i|\hat{p}_i=p) = p \quad \quad  \forall p \in [0, 1]
    \label{definition}
\end{equation}
\begin{equation}
    \hat{p}_i = \max{softmax(\vz_i)}\quad\hat{y}_i= \mathop{\arg\max}_{\{1,2, \cdots, C\}} softmax(\vz_i)
    \label{definition2}
\end{equation}
This formulation means that for example $20\%$ of all predictions with a confidence score of $80\%$ should be false.

\begin{figure*}[t]
    \centering
    \subfloat[Long-tailed distribution]{\includegraphics[width=0.3\textwidth]{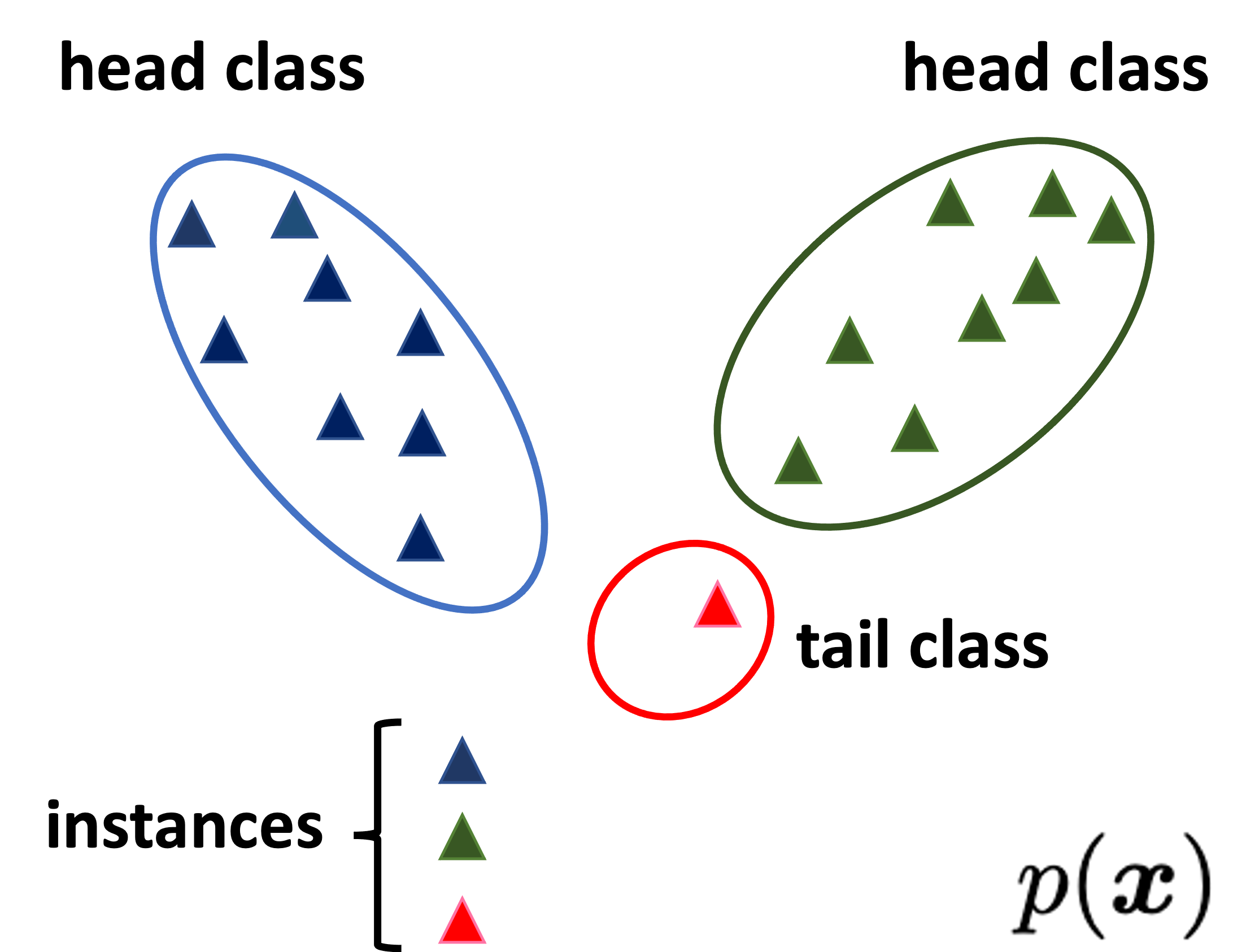}}
    \subfloat[Balanced distribution]{\includegraphics[width=0.3\textwidth]{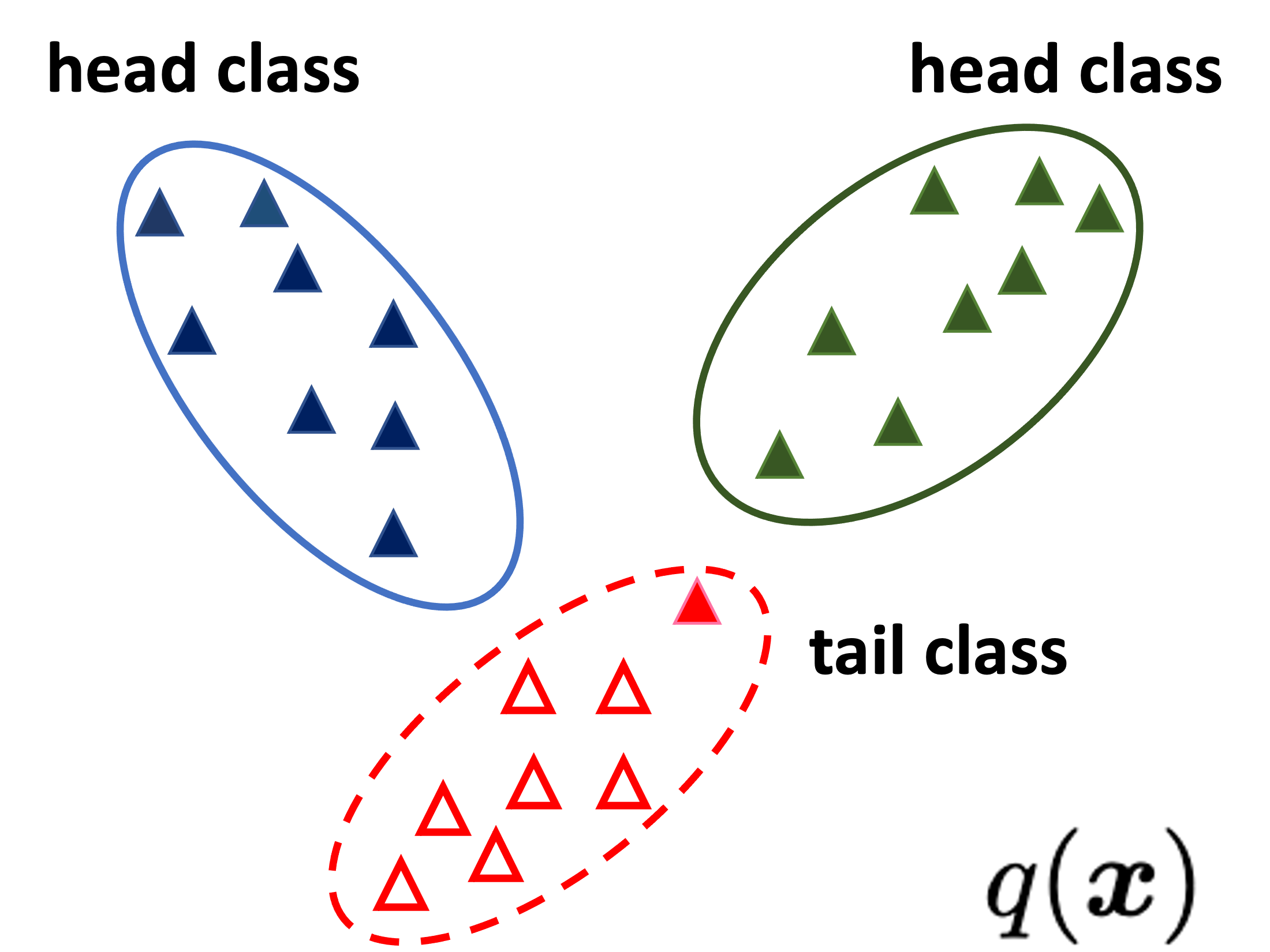}}
    \subfloat[Our estimated distribution]{\includegraphics[width=0.3\textwidth]{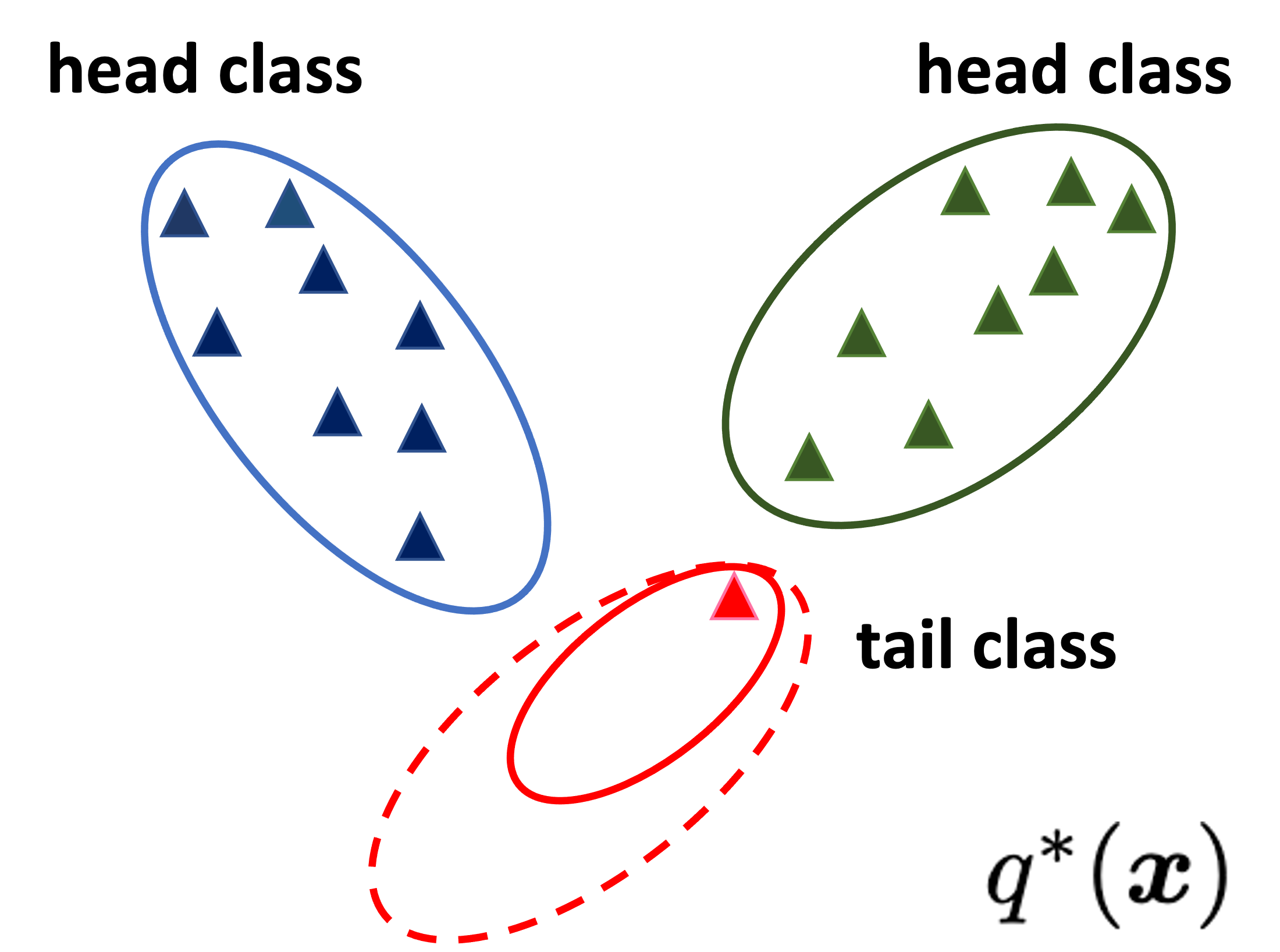}}
    \caption{(a) The long-tailed distribution $p(\vx)$. (b) The balanced distribution $q(\vx)$. Compared with (a), the distributions of head classes are the same, while the distributions of tail classes are not. (c) Our estimated distribution $q^*(\vx)$. With the help of head classes, we can estimate the distribution of tail classes and acquire their density.}
    \label{main}
    % \vspace{-0.3cm}
\end{figure*}

\textbf{Temperature scaling}. As shown in \cref{ts}, temperature scaling~\cite{guo2017calibration} fits a single parameter $T$ from the validation set and applies it to other test sets. 
\begin{equation}
    T^* = \mathop{\arg\min}_{T} \mathbb{E}_{p}[\mathcal{L}(s(\vz_i / T), y_i)] 
    % \quad \quad
    % s(\vz_i/T) = \frac{exp(\vz_i/T)}{\sum_{j=1}^Cexp(\vz_j/T)} 
    \label{ts}
\end{equation}
Similar to the training classification task, the loss function $\mathcal{L(\cdot)}$ for calibrating the temperature is the Cross Entropy loss. Since the validation set also follows long-tailed distribution while the test set does not, the learned parameter $T$ is difficult to generalize well to the test set. 

\subsection{Calibration under long-tailed distribution}
To tackle the generalization issue of the original temperature scaling in calibration under long-tailed distribution, we propose our knowledge-transferring-based temperature scaling method to achieve cross-distribution generalization. The calibration loss on the balanced target distribution $q(\vx)$ can be reformulated as the weighted calibration error of the source distribution $p(\vx)$:
\begin{equation}
    \begin{aligned}
    \mathbb{E}_{q} [\mathcal{L}(s(\vz_i / T), y_i)]&= \int_q{q(\vx_i)\mathcal{L}(s(\vz_i / T), y_i)}dx\\
    &= \int_p{\frac{q(\vx_i)}{p(\vx_i)}p(\vx_i)\mathcal{L}(s(\vz_i / T), y_i)}dx\\
    &=\mathbb{E}_{p} [w(\vx_i)\mathcal{L}(s(\vz_i / T), y_i)]
    \end{aligned}
    \label{calib}
\end{equation}
As shown in \cref{calib}, we can acquire the target distribution error $\mathbb{E}_{q} [\mathcal{L}(s(\vz_i / T), y_i)]$ by estimating the ratio of probabilities $w(\vx)=q(\vx)/p(\vx)$ for each instance. Domain adaptation calibration like TransCal~\cite{wang2020transferable} utilizes LogReg~\cite{qin1998inferences, bickel2006dirichlet} to estimate the ratio of density. It estimates the density by training a logistic regression classifier that realizes binary classification of source and target domains. Such methods cannot be used for the long-tailed calibration problem since the balanced distribution of test data is unknown and thus binary classification cannot be applied. 

We model the distributions $p(\vx)$ and $q(\vx)$ as mixtures of Gaussian distributions, respectively, by modeling each class as a Gaussian distribution. As shown in \cref{main}, because head classes have plenty of instances in both $p(\vx)$ and $q(\vx)$, the distributions of head classes in $q(\vx)$ can be viewed as the same as those in $p(\vx)$, which can be easily estimated from the training set.  However, each tail class only has a few instances in $p(\vx)$ while sufficient instances are available in $q(\vx)$, the distributions of tail classes are different in $p(\vx)$ and $q(\vx)$. Since it is difficult to acquire the balanced distribution $q(\vx)$, we constitute the estimated distribution $q^*(\vx)$ to approach the truth distribution $q(\vx)$, where the key is to approximate the probability density value of each instance in tail classes under $q^*(\vx)$ from the estimated Gaussian distributions of tail classes in $p(\vx)$. 

For each class $c \in \{1, 2, \cdots, C\}$, the corresponding distribution is $p_c(\vx) = \mathcal{N}(\vx|\vmu_{c}, \bm\Sigma_{c})$, where the mean $\vmu_{c}$ and the variance $\bm\Sigma_{c}$ are calculated by the set of output features belonging to class $c$ in the training set. When $c \in \mathcal{A}_{tail}$, its distribution $p_c(\vx)$ is not reliable due to the limited instances. It is crucial to estimate the probability density under balanced distribution. Since there exists common information between head classes and tail classes~\cite{liu2020deep, liu2021self, liu2021gistnet}, it is rational to transfer knowledge from all head classes to the $c^{th}$ tail class and recover its balanced distribution to the utmost. Normally, the more similar two classes are, the more information they share. We measure the similarity of the two classes by the Wasserstein distance to consider the first-order and second-order statistics. We construct a distance vector $\vd_c \in \mathbb{R}^{|\mathcal{A}_{head}|}$, where $\vd_c^k$ is the $k^{th}$ element of $\vd_c$ and denotes the similarity between class $c$ and head class $k$. 
\begin{equation}
    \vd_c^k = Wasserstein(p_c(\vx), p_k(\vx))
\end{equation}
Following the attention mechanism~\cite{vaswani2017attention}, we calculate the attention score $\vs_c$ from the similarities as follows,
\begin{equation}
    \vs_c = softmax(-\frac{\vd_c}{\sqrt{dim(\vf)}})
    \label{score}
\end{equation}
where the $dim()$ function acquires the feature dimension. A head class will be assigned a large attention score if its distribution is similar to the distribution of tail class $c$.

We calibrate the distribution of each tail class by transferring knowledge from all head classes based on $\vs_c$. Specifically, for the $c^{th}$ tail class, the statistics of its calibrated distribution are estimated as follows.
\begin{equation}
    \begin{aligned}
        \vmu_{c^*} &= \alpha \vmu_{c} + (1 - \alpha) \sum_{k \in \mathcal{A}_{head}} \vs_c^k\vmu_{k}\\
    \bm{\sqrt{\Sigma}}_{c^*} &= \alpha \bm{\sqrt{\Sigma}}_{c} + (1 - \alpha) \sum_{k \in \mathcal{A}_{head}} \vs_c^k\bm{\sqrt{\Sigma}}_{k}
    \end{aligned}
    \label{merge}
\end{equation}
The estimated distribution $\mathcal{N}(\vmu_{c^*}, \bm\Sigma_{c^*})$ contains the information of all head classes according to their similarity scores in $\vd_c$, where $\alpha$ is a hyper-parameter.

For each instance $\vx_i$ in tail classes, we can then acquire its probability under the estimated distribution $q_{y_i}^*(\vx_i) = \mathcal{N}(\vx_i | \vmu_{{y_i}^*}, \bm\Sigma_{{y_i}^*})$. Based on the estimated $q_{y_i}^*(\vx_i)$, the importance weight is defined in \cref{important}. 
\begin{equation}
    w^*(\vx_i) = 
    \begin{cases}
        1 &\text{$y_i \in \mathcal{A}_{head}$} \\
        min(max(\frac{q_{y_i}^*(\vx_i)}{p_{y_i}(\vx_i)}, \eta_1), \eta_2) &\text{$y_i \in \mathcal{A}_{tail}$}
    \end{cases}
    \label{important}
\end{equation}
For each instance in head classes, the importance weight equals $1$ since head classes in the two distributions are the same. For each instance in tail classes, the importance weight equals $q_{y_i}^*(\vx_i)/p_{y_i}(\vx_i)$. In practice, we restrict the value of the weight from $\eta_1$ to $\eta_2$ to avoid abnormal values. Empirically, we set $\eta_1=0.3$ and $\eta_2=5.0$.

By using the importance weight to bridge the training long-tailed distribution and the test balanced distribution, we learn the temperature $T$ in the final softmax layer on the validation set to calibrate the classification confidence. The final optimization function is shown in \cref{ours}. 
\begin{equation}
    T^* = \mathop{\arg\min}_{T} \mathbb{E}_{p}[w^*(\vx_i)\mathcal{L}(s(\vz_i / T), y_i)] 
    \label{ours}
\end{equation}

To better understand  our calibration method, we summarize the detailed procedures in Appendix 3.

\subsection{Analysis}
\newtheorem{theorem}{Theorem}[section]
\begin{theorem}
\label{coro}
We denote the distribution of the $k^{th}$ class in the long-tailed distribution, the ground truth balanced distribution, and the estimated distribution by $p_k(\vx) = \mathcal{N}(\vx|\vmu_{p_k}, \bm\Sigma_{p_k})$, $q_k(\vx) = \mathcal{N}(\vx|\vmu_{q_k}, \bm\Sigma_{q_k})$, and $q_k^*(\vx) = \mathcal{N}(\vx|\vmu_{q_k^*}, \bm\Sigma_{q_k^*})$, respectively. The absolute error $|\mathbb{E}_{p_k} [w_k(\vx)\mathcal{L}(s(\vz / T), y)] - \mathbb{E}_{p_k} [w_k^*(\vx)\mathcal{L}(s(\vz / T), y)]|$ is sensitive to $\epsilon = \mathbb{E}_{p_k} [(w_k(\vx) - w_k^*(\vx))^2]$ and the bound of $\epsilon$ is shown as follows, where the formula $d_2(\cdot||\cdot)$ presents the exponential in base 2 of the Renyi-divergence~\cite{renyi1961measures}.
\begin{equation}
   \epsilon \in [(\sqrt{d_2(q_k||p_k)} - \sqrt{d_2(q_k^*||p_k)})^2, d_2(q_k||p_k) + d_2(q_k^*||p_k)]
\end{equation}
\end{theorem}
The proof is provided in Appendix 1. \cref{coro} presents the error bound of our method, which is closely related to Renyi-divergence. It is obvious that when $q_k = q_k^*$, the lower bound reaches the minimum and equals $0$. Therefore, our estimation method aims to keep $q_k^*$ approaching $q_k$ to reduce the calibration error on the test data. %$p_k(\vx) \sim \mathcal{N}(\vmu_{p_k}, \bm\Sigma_{p_k})$ as the long-tailed distribution, $q_k(\vx) \sim \mathcal{N}(\vmu_{q_k}, \bm\Sigma_{q_k})$ as the ground truth balanced distribution, and $q_k^*(\vx) \sim \mathcal{N}(\vmu_{q_k^*}, \bm\Sigma_{q_k^*})$ as the estimated distribution, where index $k$ denotes the $k^{th}$ class. This also indicates that our estimated distribution is the same as the ground-truth distribution. 

We also explore how the importance weight influences the calibration compared with temperature scaling. For simplicity, we constitute one-dimensional Gaussian distribution $p(x) = \mathcal{N}(x|\mu_a, \sigma^2_a)$ and $q(x) = \mathcal{N}(x|\mu_b, \sigma^2_b)$, where $\mu_a \neq \mu_b$ and $\sigma^2_a < \sigma^2_b$. We draw the conclusion that the value of $w(x) > 1$ if $ \tau_1 > x$ or $x > \tau_2$, and $w(x) < 1$ if $\tau_1 <x < \tau_2$, where $\tau_1$ and $\tau_2$ are calculated as: 
\begin{equation}
    \begin{aligned}
        \Delta &= \sqrt{(\mu_a - \mu_b)^2+(\sigma^2_b-\sigma^2_a)(\ln\sigma^2_b - \ln\sigma^2_a)}\\
        \tau_1&=\frac{\mu_a\sigma^2_b - \mu_b\sigma^2_a-\sigma_a\sigma_b\Delta}{\sigma^2_b - \sigma^2_a}\\
        \tau_2&=\frac{\mu_a\sigma^2_b - \mu_b\sigma^2_a+\sigma_a\sigma_b\Delta}{\sigma^2_b - \sigma^2_a}
    \end{aligned}
\end{equation}
The proof is presented in Appendix 2. Normally, instances clustered around the mean are more likely to be classified correctly. Therefore, the instances whose importance weight $w(x) < 1$ are likely to be correctly classified while $w(x) > 1$ on the contrary. In practice, a model trained with imbalanced data generalizes well for head classes but easily overfits tail classes, and hence obtains degraded performances on balanced test data. Our importance weight estimation method assigns larger weights to instances of tail classes that are classified incorrectly.

%% file: 3.experiment.tex
\section{Experiment}
\subsection{Datasets}
\textbf{CIFAR-10-LT}. CIFAR-10-LT~\cite{cao2019learning} is simulated from balanced CIFAR-10~\cite{krizhevsky2009learning}. We conduct experiments with different imbalance factors (IF) and generate three imbalanced datasets with IF=100, IF=50, and IF=10, respectively. For each dataset, we randomly split $80\%$ instances as the training set and $20\%$ as the validation set. For comparison, we use four test sets: (1) original CIFAR-10 test set, (2) CIFAR-10.1~\cite{recht2018cifar}, (3) CIFAR10.1-C~\cite{hendrycks2019robustness}: 95 synthetics datasets generated on CIFAR-10.1 with different transformations, (4) CIFAR-F~\cite{sun2021label}: 20 real-word test sets collected from Flickr. \textbf{MNIST-LT}. MNIST-LT is simulated from MNIST~\cite{lecun1998gradient}. Similar to CIFAR-10-LT, we generate three imbalanced datasets with IF=100, IF=50, and IF=10, respectively. For comparison, we use four test sets: (1) original MNIST test set, (2) SVHN~\cite{netzer2011reading}, (3) USPS~\cite{hull1994database}, (4) Digital-S~\cite{sun2021label}: 5 test sets that are searched from Shutterstock based on different options of color. Note that the original MNIST test set is slightly imbalanced, which is closer to reality. \textbf{CIFAR-100-LT}. CIFAR-100-LT~\cite{cao2019learning} is generated from the CIFAR-100 dataset. We generate imbalanced datasets with IF=10 and conduct experiments on the original CIFAR-100 test set. \textbf{ImageNet-LT}. ImageNet-LT~\cite{liu2019large} is simulated from ImageNet~\cite{deng2009imagenet}. We merge the long-tailed training set and balanced validation set from the original ImageNet-LT. Following the principle of CIFAR-10-LT, we generate a long-tailed training set and a long-tailed validation set. We conduct experiments on a balanced test set. More details are presented in Appendix 4.

\begin{table*}[h]
\centering
\small
\begin{tabular}{ll|lllllllll}
\toprule
\multirow{2}{*}{\bf{IF}}&\multirow{2}{*}{\bf{Dataset}}  &\multicolumn{8}{c}{\bf{Method}}
\\
 & &Base &TS &ETS &TS-IR &IR & IROvA &SBC & GPC& Ours\\
  \midrule
\multirow{4}{*}{IF=100}&
CIFAR-10 &21.79 &12.24 &12.16 &11.64 &12.36 &13.36 &12.13 &11.65 &\textbf{9.84}\\
&CIFAR-10.1 &28.97 &16.75 &16.70 &16.65 &17.13 &17.93 &16.78 &15.71 &\textbf{13.86}\\
&CIFAR-10.1-C &58.22 &43.01 &43.00 &43.05 &43.34 &43.83 &42.53 &41.98 &\textbf{39.58} \\
&CIFAR-F &29.22 &15.27 &15.24 &15.52 &15.75 &16.23 &15.45 &14.18 &\textbf{12.15}\\
\hline
\multirow{4}{*}{IF=50}&
CIFAR-10 &17.36 &7.65 &8.04 &8.22 &9.75 &9.45 &7.55 &7.78 & \textbf{3.99}\\
&CIFAR-10.1 &22.79 &10.36 &10.99 &11.72 &13.35 &12.70 &10.32 &10.82 &\textbf{5.74}\\
&CIFAR-10.1-C &55.52 &38.66 &39.9 &40.16 &41.58 &40.76 &38.94 &39.39 &\textbf{33.09} \\
&CIFAR-F &25.37 &11.30 &12.21 &12.67 &14.39 &13.37 &11.4 &11.76 &\textbf{6.64}\\
\hline
\multirow{4}{*}{IF=10}&
CIFAR-10 &8.39 &2.23 &1.64 &2.03 &2.29 &2.42 &2.49 &2.01 & \textbf{1.00}\\
&CIFAR-10.1 &13.80 &4.87 &4.25 &4.54 &5.38 &5.23 &5.63 &4.66 &\textbf{3.95}\\
&CIFAR-10.1-C &48.31 &32.77 &31.07 &32.11 &32.29 &31.94 &33.16 &31.37 &\textbf{29.98} \\
&CIFAR-F &19.73 &8.15 &6.80 &8.42 &8.97 &8.13 &8.54 &7.10 &\textbf{5.97}\\
\bottomrule
\end{tabular}
\caption{The ECE (\%) on CIFAR-10-LT.}
\label{CIFAR10}
\end{table*}

\begin{table*}[h]
\centering
\small
\begin{tabular}{ll|lllllllll}
\toprule
\multirow{2}{*}{\bf{IF}}&\multirow{2}{*}{\bf{Dataset}}  &\multicolumn{8}{c}{\bf{Method}}
\\
 & &Base &TS &ETS &TS-IR &IR & IROvA &SBC &GPC &Ours\\
  \midrule
\multirow{4}{*}{IF=100}&
MNIST &2.52 &1.27 &1.84 &2.82 &2.84 &1.84 &1.92 &1.76 &\textbf{1.08}\\
&SVHN &16.06 &7.20 &11.62 &21.25 &22.18 &14.93 &9.59 &13.67 &\textbf{6.09}\\
&USPS &15.00 &9.52 &12.25 &13.25 &13.62 &10.58 &10.10 &11.44 &\textbf{8.40} \\
&Digital-S &32.10 &22.13 &27.35 &30.13 &31.01 &27.48 &23.34 &27.60 &\textbf{20.28}\\
\hline
\multirow{4}{*}{IF=50}&
MNIST &1.12 &0.85 &1.14 &1.53 &1.54 &1.02 &1.01 &1.12& \textbf{0.79}\\
&SVHN &\textbf{2.32} &3.95 &3.33 &11.42 &12.15 &2.63 &9.43 &\textbf{2.32} &4.53\\
&USPS &11.21 &8.14 &12.81 &11.89 &11.91 &10.54 &8.57  &11.21 &\textbf{8.02} \\
&Digital-S &15.22 &10.81 &17.81 &20.96 &21.81 &13.64 &16.74 &15.18 &\textbf{10.34}\\
\hline
\multirow{4}{*}{IF=10}&
MNIST &0.56 &0.23 &\textbf{0.21} &0.50 &0.52 &0.23 &0.25 &0.41 &0.36\\
&SVHN &5.75 &6.76 &6.94 &8.10 &\textbf{4.51} &5.31 &7.00 &5.31 &7.43\\
&USPS &8.29 &4.81 &4.60 &6.59 &6.98 &4.76 &5.12  &5.88 &\textbf{4.55} \\
&Digital-S &13.55 &8.21 &8.09 &15.37 &13.34 &8.31 &7.67 &8.24&\textbf{7.37}\\
\bottomrule
\end{tabular}
\caption{The ECE (\%) on MNIST-LT.}
\label{mnist}
\end{table*}

\begin{table*}[h]
\centering
\small
\begin{tabular}{ll|lllllllll}
\toprule
\multirow{2}{*}{\bf{Model}}&\multirow{2}{*}{\bf{Dataset}}  &\multicolumn{8}{c}{\bf{Method}}\\
 & &Base &TS &ETS &TS-IR &IR & IROvA &SBC &GPC &Ours\\
  \midrule
ResNet-32& CIFAR-100 &20.38 &2.50 &2.10 &6.07 &9.35 &5.92 &6.74 &3.27& \textbf{1.50}\\
DenseNet-40& CIFAR-100 &16.00 &3.43 &2.51 &5.57 &8.42 &5.76 &5.96 &2.73 &\textbf{2.37}\\
VGG-19& CIFAR-100 &27.86 &3.81 &2.36 &6.35 &10.35 &6.66 &8.03 &3.82 &\textbf{1.99} \\
\bottomrule
\end{tabular}
\caption{The ECE (\%) on CIFAR-100-LT.}
\label{cifar100}
\end{table*}

\begin{table*}[!h]
\centering
\small
\begin{tabular}{ll|lllllllll}
\toprule
\multirow{2}{*}{\bf{Model}}&\multirow{2}{*}{\bf{Dataset}}  &\multicolumn{8}{c}{\bf{Method}}\\
 & &Base &TS &ETS &TS-IR &IR & IROvA& SBC &GPC & Ours\\
  \midrule
ResNet-50& ImageNet &10.18 &6.72 &6.06 &10.23 &11.15 &7.63 &9.12 &5.46 & \textbf{3.45}\\
\bottomrule
\end{tabular}
\caption{The ECE (\%) on ImageNet-LT.}
\label{imagenet}
\end{table*}

\subsection{Experiments set up}
\textbf{Classification model}. We use our method to calibrate different classification models. For CIFAR-10-LT and MNIST-LT, we use ResNet-32~\cite{he2016deep} and LeNet-5~\cite{lecun1998gradient} as the classification model, respectively. To verify our method can be applied to different models, we apply ResNet-32, DenseNet-40~\cite{huang2017densely}, VGG-19~\cite{simonyan2014very} as classification models and test on CIFAR-100-LT, respectively. We do experiments on the large-scale dataset, ImageNet-LT, with ResNet-50. Details about training strategies are presented in Appendix 4. \textbf{Metrics/Baselines}. The most popular evaluation metric for calibration is ECE~\cite{naeini2015obtaining}. Besides, we also use SCE~\cite{nixon2019measuring} and ACE~\cite{neumann2018relaxed} as evaluation metrics. We compare our method with temperature scaling (TS)~\cite{guo2017calibration}, ETS~\cite{zhang2020mix}, TS-IR~\cite{zhang2020mix}, IROvA~\cite{zadrozny2002transforming}, IRM~\cite{zhang2020mix},  Scaling-binning calibrator (SBC)~\cite{kumar2019verified} and GPC~\cite{wenger2020non}. As for our method, all the experiments are conducted with hyper-parameter $\alpha=0.998$ if not specified. Code is available: \url{https://github.com/JiahaoChen1/Calibration}

\subsection{Results}
\textbf{CIFAR-10-LT}. As shown in \cref{CIFAR10}, our method achieves the best performance on the CIFAR-10-LT dataset. Usually, the model trained with a heavier imbalanced dataset suffers heavier miscalibration. Our method can realize competitive results in different situations. Since CIFAR-10.1 and CIFAR-F are collected from the real-world, the excellent results indicate that our method can generalize to different domains. The results of CIFAR-10.1-C also verify that our method can handle datasets containing different transformations. \textbf{MNIST-LT}. \cref{mnist} demonstrates the effectiveness of our method on MNIST-LT. Except for (IF=50: SVHN) and (IF=10: MNIST, SVHN), our method achieves the best performance. Although our method obtains negative optimization on (IF=50: SVHN) and (IF=10: SVHN), and lower performance compared with ETS on (IF=10: MNIST), the performance of our method is still acceptable. Overall, our method outperforms other methods significantly in most cases. \textbf{CIFAR-100-LT}. For the CIFAR-100-LT benchmark, our method achieves the best results on calibrating different models as shown in \cref{cifar100}. We do experiments on three different architectures including ResNet, DenseNet, and VGG. Compared with DenseNet, the performance gains of ResNet and VGG are even higher, and our method achieves the smallest ECE on ResNet. \textbf{ImageNet-LT}. As shown in \cref{imagenet}, our method achieves the best performance on the ImageNet-LT benchmark. This indicates that our method can be scaled to large-scale datasets. For detail, our method reduces the ECE value from $10.18\%$ to $3.45\%$ while the second best method, GPC, can only reduce it to $5.46\%$. For all datasets, the results of SCE and ACE are presented in Appendix 6 and similar conclusions can be observed.

\begin{figure}[t]
    \centering
    \subfloat[$\alpha=0.998$]{\includegraphics[width=0.16\textwidth]{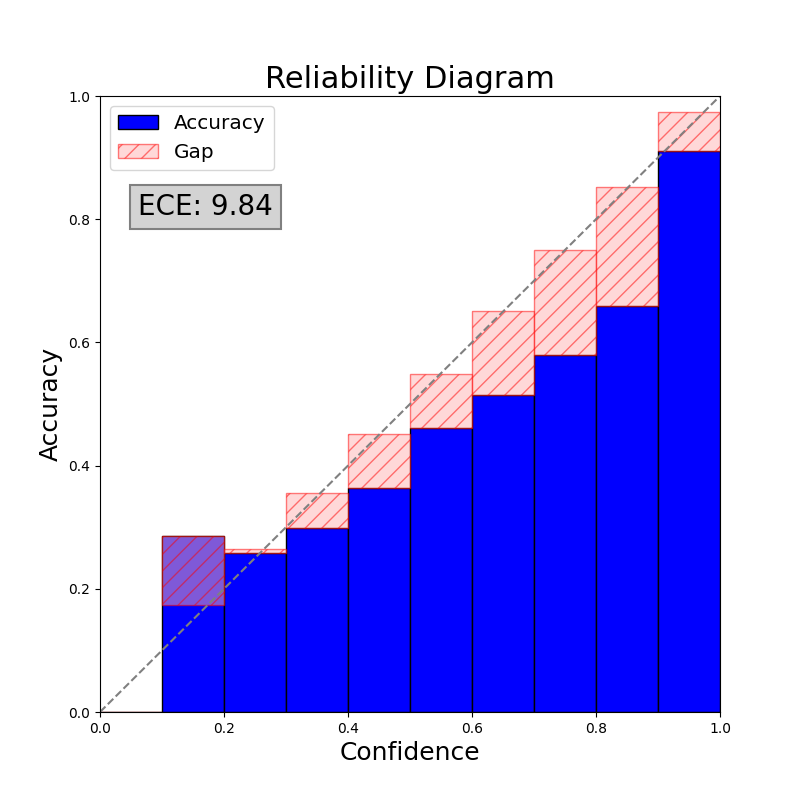}}
     \subfloat[$\alpha=0.997$]{\includegraphics[width=0.16\textwidth]{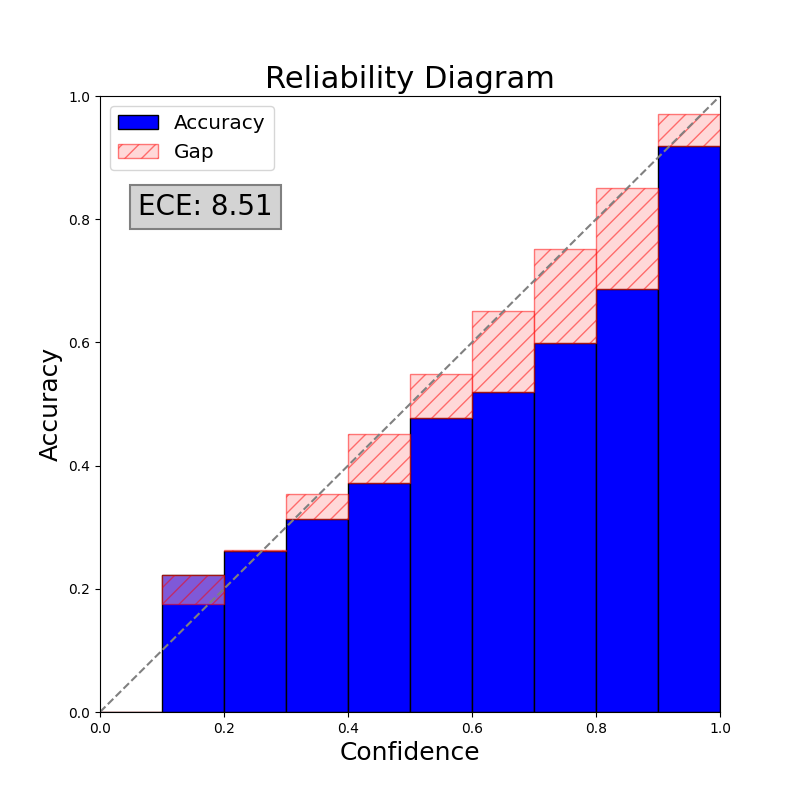}}
    \subfloat[$\alpha=0.996$]{\includegraphics[width=0.16\textwidth]{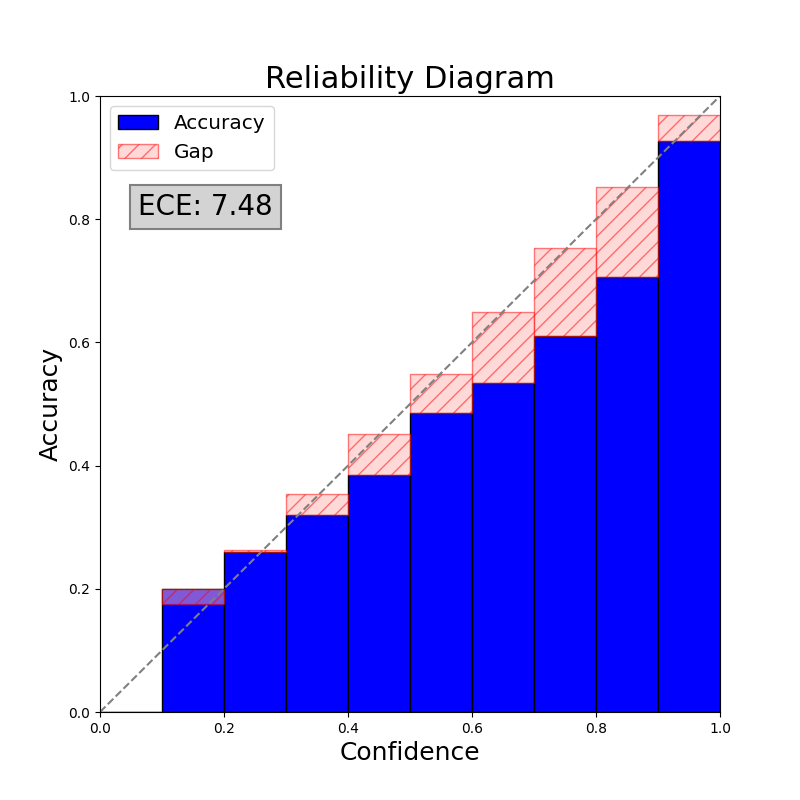}}
    \caption{The reliability diagram of our method with (a) $\alpha=0.998$, (b) $\alpha=0.997$, and (c)  $\alpha=0.996$. }
    \label{diagram}
    \vskip -0.1in
\end{figure}

\subsection{Ablation study}
\footnotetext{More ablation studies are in Appendix 5}
\textbf{Reliability diagram}. We also visualize the reliability diagram of our method with different $\alpha$ on the CIFAR-10 test set. As shown in \cref{diagram}, our method can give more reliable results and alleviate the overconfidence problem. The reliability diagrams of the baseline and TS are shown in \cref{phenomenon}, and our method achieves competitive results compared with them. Specifically, a higher value of $\alpha$ achieves better results in the reliability diagram on this benchmark. Compared with TS, our method well calibrates the predictions for instances with high confidence. We also observe that our method leads to slight underconfidence in instances with very small confidence values, this may be because these samples themselves are more difficult to classify. %, the overall performance is acceptable.

\begin{figure}[t]
    \centering
    \includegraphics[width=0.35\textwidth]{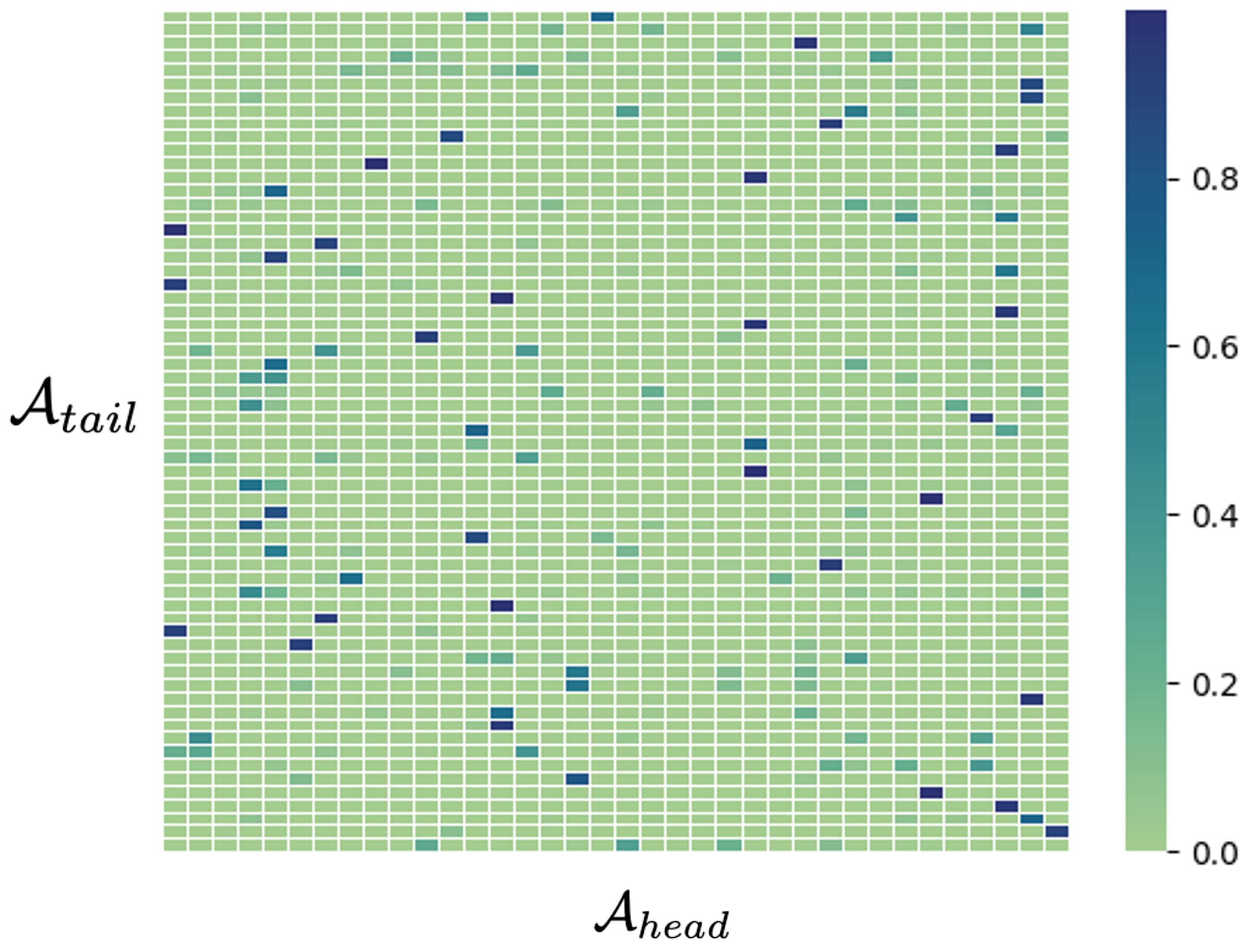}
    \caption{Experiment on CIFAR-100-LT. The shape of the matrix is $|\mathcal{A}_{tail}| \times |\mathcal{A}_{head}|$. Each row denotes the attention vector $\vs$.}
    \label{heatmap}
    \vskip -0.1in
\end{figure}

\begin{figure*}[h]
    \centering
    \subfloat[IF=100]{\includegraphics[width=0.3\textwidth]{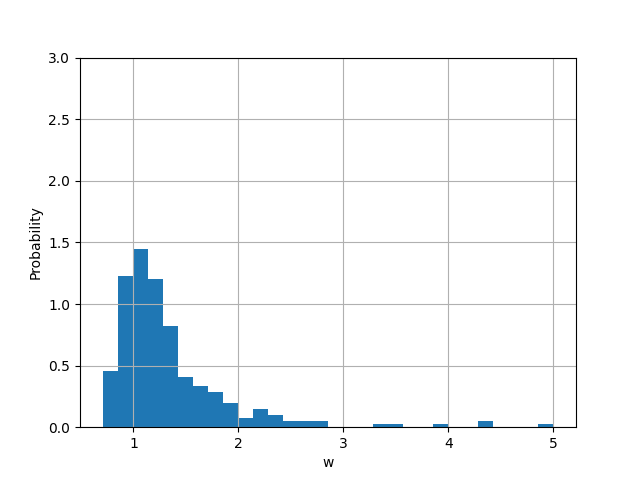}}
     \subfloat[IF=50]{\includegraphics[width=0.3\textwidth]{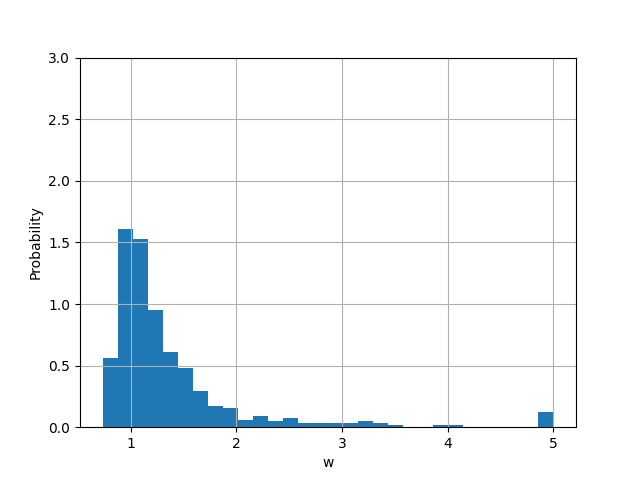}}
    \subfloat[IF=10]{\includegraphics[width=0.3\textwidth]{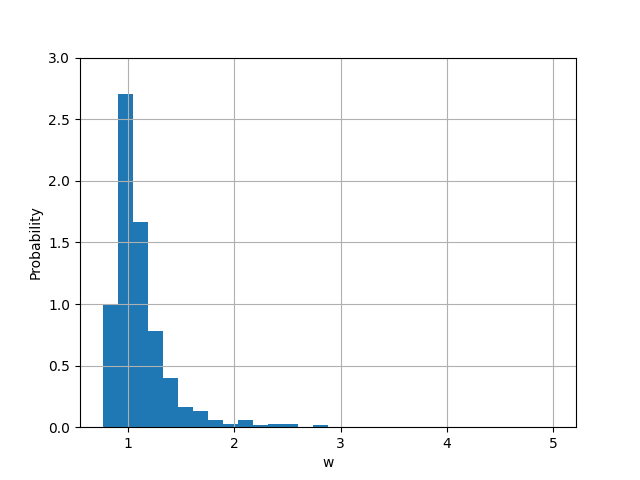}}
    \caption{The distribution histograms of $w^*(\vx)$ with $\alpha=0.998$. The horizontal axis represents the value of $w^*(\vx)$ and the vertical axis represents the probability density. The model is trained on (a) CIFAR-10-LT with IF=100, (b) CIFAR-10-LT with IF=50, and (c) CIFAR-10-LT with IF=10, respectively.}
    \label{wx}
    \vskip -0.1in
\end{figure*}

\begin{figure*}
	\begin{minipage}[h]{0.57\linewidth}
		\centering
		\subfloat[CIFAR-10 test set]{\includegraphics[width=0.5\textwidth]{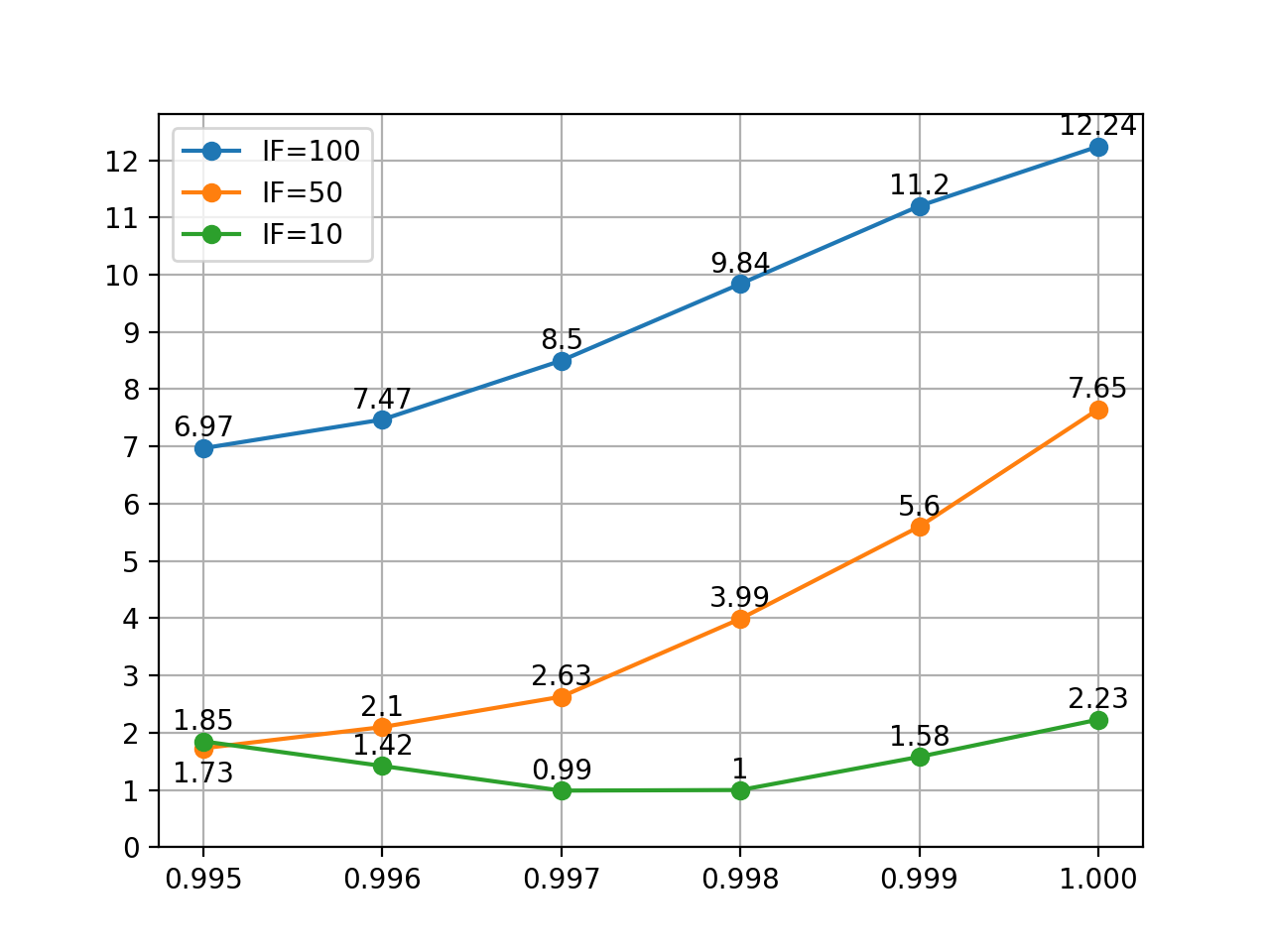}}
    \subfloat[The value of temperature]{\includegraphics[width=0.5\textwidth]{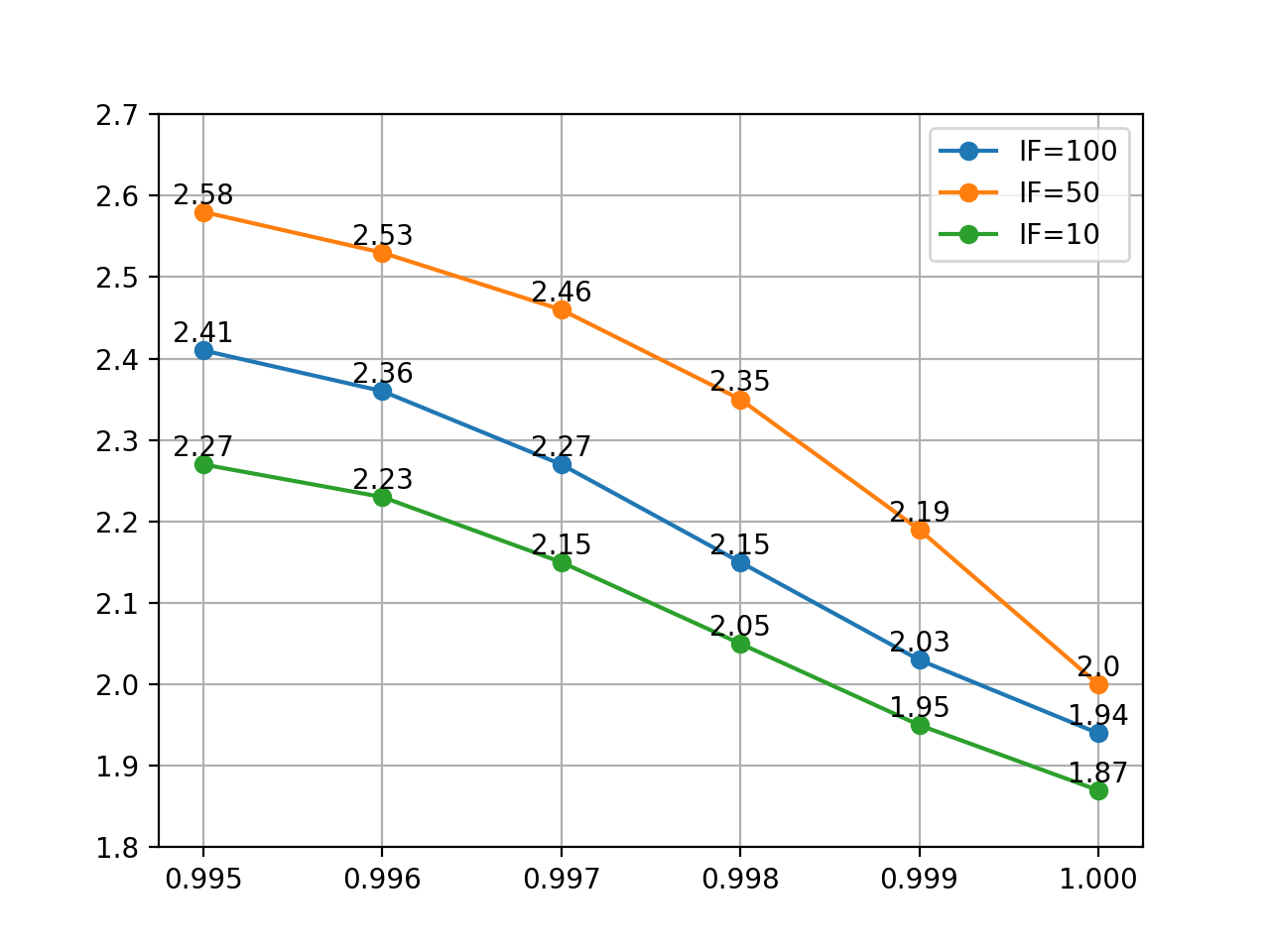}}
    \caption{The blue line, orange line, and green line denote results on CIFAR-10-LT with IF=100, IF=50, and IF=10, respectively. The horizontal axis represents the hyper-parameter $\alpha$. (a) The vertical axis represents the ECE value, which is tested on the original CIFAR-10 test set. (b) The vertical axis represents the temperature value, which is tested on the original CIFAR-10 test set.}
    \label{abla}
	\end{minipage}
 \quad
        \begin{minipage}[h]{0.41\linewidth}
		\centering
		\begin{tabular}{l|lll}
    \toprule
    \bf{Dataset} & \bf{Uniform} & \bf{OneHot} & \bf{Ours}\\
    \midrule
    CIFAR-10        &8.10 &7.42  &6.97\\
    CIFAR-10.1      &11.72 &10.91  &10.4\\
    CIFAR-10.1-C    &37.09 &36.17  &35.59\\
    CIFAR-F         &9.91 &9.02  &8.46\\
    \bottomrule
    \end{tabular}
    % \captionof{table}{Table}
    \captionof{table}{The ECE ($\%$) on CIFAR-10-LT with IF=100, $\alpha$=0.995. Uniform denotes that we transfer knowledge from all head classes equally, i.e., all head classes give the same contribution. OneHot denotes that $\vs$ is a one-hot attention vector and the weight of the most similar head class is $1$ while others are $0$.}
    \label{strategy}
	\end{minipage}
 \vskip -0.15in
\end{figure*}

\textbf{The visualization of attentions}. Our method utilizes the distance between different distributions and acquires the attention score by the softmax function. As shown in \cref{heatmap}, each tail class has multiple similar head classes and their knowledge will be used. For example, the top two similar classes of tail class ``woman" are ``girl" and ``boy", while the not similar classes are ``cloud" and ``castle". It is obvious that ``woman" has much similar information with ``girl" and ``boy" and we utilize more information from such classes is rational. In addition, we do experiments with two different knowledge-transferring strategies. One strategy is that we only transfer knowledge from the most similar head class and discard other head classes, which is called OneHot. The other strategy is that we transfer knowledge from all head classes equally and ignore their inherent differences, which is called Uniform. The results in \cref{strategy} show the utility of our strategy and prove the importance to consider the knowledge of all head classes according to their relevance.

\textbf{The distribution of $w^*(\vx)$}. Our method is heavily influenced by the value of $w^*(\vx)$, so we explore the distribution of $w^*(\vx)$ on different datasets. We do experiments on CIFAR-10-LT with IF=100, IF=50, and IF=10, respectively. As shown in \cref{wx}, the overall distribution of $w$ is clustered around the value $w=1$. The more imbalanced the dataset, the more instances with larger $w$ values. More instances have larger values of $w$ at IF=100 than at IF=10. Since the dataset with IF=100 suffers a heavier imbalance, it faces a more serious domain shift, and more instances equipped with a large value of $w$ are rational.%Compared with the IF=10 dataset, more instances have a larger value of $w$ on the IF=100 dataset.

\textbf{Ablation study on hyper-parameter $\alpha$}. The most important hyper-parameter of our method is $\alpha$, which controls how much the information of the head class is transferred. Normally, a smaller value of $\alpha$ means we utilize more information from head classes. As shown in \cref{abla}, with the growth of value $\alpha$, the value of our temperature exhibits a downtrend. Note that $\alpha=1.0$ represent traditional temperature scaling since $w(\vx) = 1$ for all instances. A larger temperature can relieve the overconfidence phenomenon effectively. In addition, a heavily imbalanced dataset (IF=100) needs a larger temperature value compared with a slightly imbalanced dataset (IF=10).

As shown in \cref{abla}, for the performance on the CIFAR-10 test set, different $\alpha$ achieve different performances. Actually, a heavily imbalanced dataset (IF=100) achieves the best performance on $\alpha=0.995$ while a slightly imbalanced dataset (IF=10) achieves the best on $\alpha=0.997$. This indicates that we need to utilize more information from head classes if fewer instances of tail classes are available.  It is interesting that there is a downtrend and uptrend in the green curve in \cref{abla} (a). The reason is that the model is underconfident when $\alpha < 0.997$ because of the larger value of temperature. Therefore, it is not a good choice to calibrate every model with a small value of $\alpha$ and it is proper to apply a smaller value of $\alpha$ on a heavily imbalanced dataset and a larger value of $\alpha$ on a slightly imbalanced dataset.

% \begin{table}[]
%     \centering
%     \begin{tabular}{l|lll}
%     \toprule
%     \bf{Dataset} & \bf{Uniform} & \bf{OneHot} & \bf{Ours}\\
%     \midrule
%     CIFAR-10        &8.10 &7.42  &6.97\\
%     CIFAR-10.1      &11.72 &10.91  &10.4\\
%     CIFAR-10.1-C    &37.09 &36.17  &35.59\\
%     CIFAR-F         &9.91 &9.02  &8.46\\
%     \bottomrule
%     \end{tabular}
%     \caption{The ablation study of $\vs$ on CIFAR-10-LT with IF=100. Uniform denotes we treat all head classes equally and they have the same weights. OneHot denotes $\vs$ as a one-hot vector and the weight of the most similar head class is $1$ while others are $0$. All experiments are conducted with $\alpha$=0.995}
%     \label{tab:my_label}
% \end{table}

% However, the CIFAR-10.1-C dataset presents different results and we achieve the best performance with $\alpha=0.995$ for all imbalanced situations. Since CIFAR-10.1-C are synthetic datasets generated from CIFAR-10.1, it suffers a heavier domain shift compared with the CIFAR-10 test set and needs a larger value of temperature to relieve overconfidence.

%% file: 4.conclusion.tex
\section{Conclusion}
% In this paper, we propose a novel importance weight-based strategy to realize post-processing calibration under long-tailed distribution. Different from traditional calibration tasks, the tackled problem faces the challenge that the validation set follows a long-tailed distribution while the distribution of the test data is balanced. To this end, we utilize the importance weight strategy to re-weight instances of tail classes. Since it is difficult to acquire the target probability density, we model the distribution of each class as a Gaussian distribution and enhance the estimation of tail class distributions by transferring knowledge from head classes. Extensive experiments on four benchmarks show the effectiveness of our method. In our future work, we intend to explore regularization terms to compensate for the imbalanced influences of head and tail classes for training calibrated models under the long-tailed distribution.
In this paper, we propose a novel importance weight-based strategy to achieve post-processing calibration under long-tailed distribution. The tackled problem differs from traditional calibration tasks as the validation set follows a long-tailed distribution, while the test data distribution is balanced. We use the importance weight strategy to re-weight instances of tail classes and model the distribution of each class as a Gaussian distribution. We enhance the estimation of tail class distributions by transferring knowledge from head classes. Extensive experiments on four benchmarks demonstrate the effectiveness of our method.

\section*{Acknowledgements}
This work was supported in part by the National Natural Science Foundation of China No. 61976206 and No. 61832017, Beijing Outstanding Young Scientist Program NO. BJJWZYJH012019100020098, Beijing Academy of Artificial Intelligence (BAAI), the Fundamental Research Funds for the Central Universities, the Research Funds of Renmin University of China 21XNLG05, and Public Computing Cloud, Renmin University of China.